# Realization of Stochastic Neural Networks and Its Potential Applications.

Sadra Rahimi Kari
*Department of Electrical and Computer engineering,*
*Istanbul Technical University*
Istanbul, Turkey
https://orcid.org/0000-0001-8230-2165

*Abstract*— Successive Cancellation Decoders have come a long way since the implementation of traditional SC decoders, but there still is a potential for improvement. The main struggle over the years was to find an optimal algorithm to implement them. Most of the proposed algorithms are not practical enough to be implemented in real-life. In this research, we aim to introduce the Efficiency of stochastic neural networks as an SC decoder and Find the possible ways of improving its performance and practicality. In this paper, after a brief introduction to stochastic neurons and SNNs, we introduce methods to realize Stochastic NNs on both deterministic and stochastic platforms.

*Keywords— Stochastic Neural Networks, Magnetic Tunnel Junctions, Spiking Neural Networks, Polar codes.*

## I. Introduction

Over the years, there has been a significant increase in the popularity and utilization of digital communications. Digital communications use a sequence of binary bits as information. To send these digital pieces of information through communicational channels, first, these binary data should be converted to the analog signal forms, then using communicational channels, these analog data is transmitted through the channel, to the receiver. At the receiver, the incoming data is converted back into digital/binary forms. The problem in this process is the modulated noise on the original signal, which can cause errors and faults in the received information/data. To detect and correct any potential errors caused by communicational channels, channel coding techniques have been introduced.

All of these methods encode information bits in different algorithms so that after receiving them, they could be recovered and corrected by decoding them into the original form. Many popular coding techniques have been proposed over the years, one of the most promising coding techniques is called Polar Codes, introduced first in 2009 by Arikan in [5]. Polar Codes are a recent breakthrough in coding theory owing to the fact that they represent a class of error-correcting codes that is mathematically proven to achieve channel capacity with low complexity encoding and decoding algorithms [7]. Since Polar Codes' introduction, they have shown high promising potential in the design of next-gen communication technologies. Nowadays, Polar Codes are used for different applications under 5G technology such as Enhanced Mobile Broadband (eMBB), Ultra Low-latency Communications (URLLC), and Massive Machine Type Communications (mMTC) [8]. One of the challenges in this area is to find the best algorithms to achieve high performances. The practical implementation of polar codes has not been investigated completely and there are only a few proposed algorithms so far. In this research, we aim to further investigate the potential of the Stochastic Neural Networks, on enhancing the performance of the Successive Cancellation decoders.

Stochastic computation, exploits the statistical nature of application-level performance metrics of emerging applications, and matches it to the statistical attributes of the underlying device and circuit fabrics [9]. There are many applications introduced for stochastic computing over the years. A particularly interesting application of stochastic computing is in the field of error correction decoding [1], and Stochastic Neural Networks (Spiking Neural Networks).

The outline of the paper is as follows. Section II is an introduction of stochastic neurons. We will review the application of Magnetic tunnel Junction devices to use them as stochastic neurons and synapses. In section III, we will introduce Stochastic Neural Networks by the realization of Spiking neural networks. Also, it is worth mentioning that changing the device structure for some practical applications may not be the best option. For this purpose, we will introduce SGD as an optimization method for ANNs. Section IV is dedicated to the conclusion, which concludes this article with insights and future directions.

## II. STOCHASTIC NEURONS

Before introducing stochastic behaviors in modeling neurons, we should first address the function of the Artificial neuron compared with the biological function of Biological neurons and synapses in our nervous system. Fig. 1 shows a biological neuron and an Artificial neuron. By definition, Artificial Neuronal Network is a program based on a very simplified algorithm of biological neural networks. These ANN are designed to mimic the logic of how the human brain thinks. We should consider two facts in comparison of artificial NNs and biological NNs:

1. Number of synapses is about 10,000 times the number of neurons. If we are trying to simulate the biological NNs using ANNs, we should consider the fact that even simulating one neuron and a corresponding synaptic system requires very large area sizes. We need to use memory compression techniques.

2. Architecture that we use in implementing ANNs is based on Von Neumann's architecture in which memory units and processing units are separated, and they communicate with each other using a data bus line. This design will lead to a memory bottleneck problem. To solve this issue, some techniques such as computing-in-memory can help.

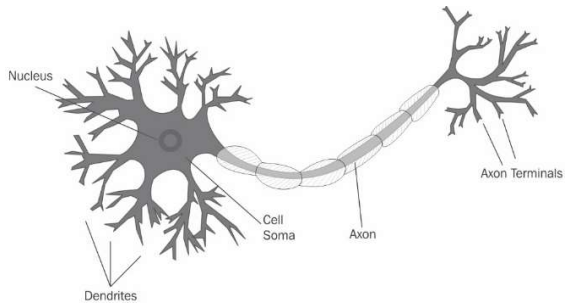 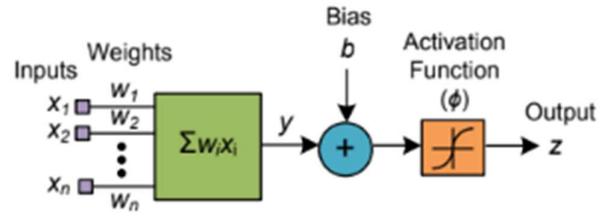

Fig.1. Biological Neuron(Left) and artificial neuron (Right).

Inefficiency in terms of energy consumption and processing power is the real issue in using ANN and simulation models designed to mimic the function of the brain. Implementing ANNs using conventional architecture (Von Neumann's architecture based on deterministic computing) has proven to be inefficient. Now we summarize the reasons for this inefficiency:

- Currently, we don't have complete knowledge over biological neurons algorithm. So, we are using simplified algorithms to mimic the overall function of a biological neuron. We need to reduce the energy gap between the brain and ANN by achieving bio-possible algorithms. This fact also means, we need more research on brain activity.

- Computing architecture of the brain possibly is not Like von Neumann's architecture (we have limited knowledge of brain activity).

- We are implementing very basic algorithmic Models based on brain activity on a platform other than a biological neural network. For example, we are using CMOS technology in our designs.

Based on the above arguments, we can understand that new architecture and design technology are needed to implement more efficient and more powerful NNs.

Stochastic computing is an emerging computation manner which is considered to be promising for efficient probability computing. Compared with the conventional computation approach with deterministic manner, stochastic computing has smaller hardware footprint and higher energy efficiency because the basic multiplication and accumulation operations can be executed by AND gate and multiplexer (MUX), respectively. As a result, very large-scale integration on one chip is much easier to realize using stochastic computing than conventional fixed-point hardware [1]. So, with the definition of stochastic computing, we need to utilize this method in order to achieve better energy efficiency and small area sizes.

In order to reach this goal, we need to utilize stochastic neural networks. The interconnection of neurons with each other from different layers creates a neural network.

Using stochastic bits as neurons and synapses, we can create Stochastic NNs. The basic element for Synapses and neurons in this study is going to be stochastic neuron implemented by Magnetic Tunnel Junctions (MTJs).

The stochastic switching of spintronic devices, such as magnetic tunnel junction (MTJ) provides a high-quality entropy source for RNG (Random Number Generators). Based on intrinsically unpredictable physical phenomenon, it can supply real random bitstreams by special circuit designs. Moreover, MTJ is promising to operate normally in sub 0.1 V, which is another bottleneck of semiconductor devices. Compared with the conventional CMOS based RNGs, the MTJ based circuit design can effectively achieve simplified structure, more compact area, higher speed and better energy-efficiency [1].

MTJ consists of one oxide barrier sandwiched by two ferromagnetic layers. The resistance of MTJ depends on the relative magnetization orientation of the two FM layers (RP at parallel (P) state and RAP at antiparallel (AP) state). Fig. 2 shows a typical structure of a MTJ device [1].

### A) Stochastic switching of MTJs:

If we are applying a current pulse into the MTJ device, based on the magnitude of the current, there is a probability of switching associated with it. Also, this switching probability depends on device properties. As shown in Fig. 3 (a), activation (switching) probability of MTJ based on input current creates a sigmoid like function. We refer to this as Stochastic Sigmoid [2].

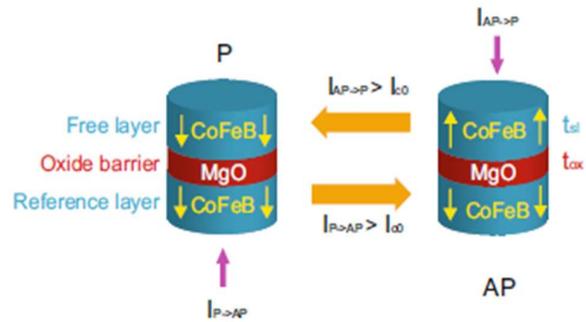

Fig.2. Typical structure of the MTJ device, which mainly consists of three layers: two FM layers separated by an insulating oxide barrier (Magnesium Oxide).

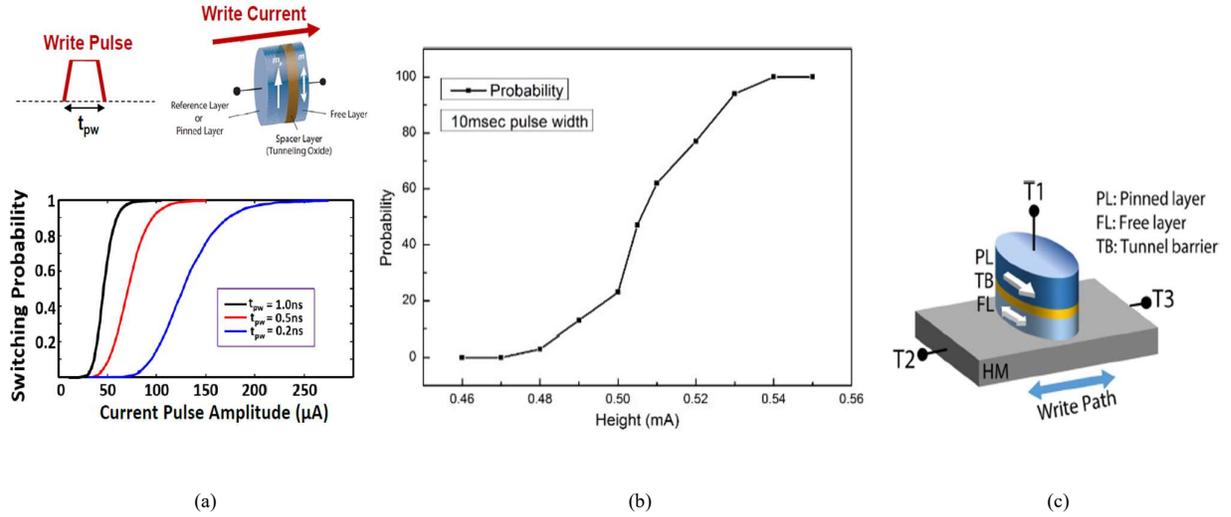

Fig.3. a) Switching probability of MTJ (without HM), b) switching probability of enhanced device, c) basic structure of MTJ and HM.

Using MTJ as a switch is inefficient because writing is a difficult process. The solution to this problem is Using heavy metals such as platinum or Tantalum. Switching, in this case, is not through the Magnetic Tunnel Junction. The stochastic spin device that will be primarily considered in this paper is shown in Fig. 3 (b). The thermal noise can be theoretically modeled as an additional thermal field,

$$\mathbf{H}_{thermal} = \sqrt{\frac{\alpha}{1+\alpha^2}\frac{2k_BT}{\gamma\mu_0 M_s V \Delta t}} G_0$$

(where G0,1 is a Gaussian distribution with zero mean and unit standard deviation, kB is the Boltzmann constant, T is the temperature, and Dt is the simulation time-step) in the Landau-Lifshitz-Gilbert (LLG) magnetization dynamic equation given by:

$$\frac{d\hat{\mathbf{m}}}{dt} = -\gamma(\hat{\mathbf{m}} \times \mathbf{H}_{eff}) + \alpha\left(\hat{\mathbf{m}} \times \frac{d\hat{\mathbf{m}}}{dt}\right) + \frac{1}{qN_s}(\hat{\mathbf{m}} \times \mathbf{I}_s \times \hat{\mathbf{m}}),$$

where $\hat{m}$ is the unit vector of "free" layer magnetization, $\gamma = \frac{2\mu_B\mu_0}{\hbar}$ is the gyromagnetic ratio for the electron, a is Gilbert's damping ratio, $H_{eff}$ is the effective magnetic field, $N_s = \frac{M_s V}{\mu_B}$ is the number of spins in the free layer of volume V ($M_s$ is the saturation magnetization and $\mu_B$ is the Bohr magneton), and Is $I_s$ the input spin current [3].

After defining the functions and basic properties of MTJ with HM, now it is time to define stochastic neurons using MTJ.

First, we need to take a look at how Biological Neurons and Artificial neurons work. Fig. 4 shows a basic Artificial Neuron block with corresponding inputs.

In the simple ANN model, Input spikes (weighted summations) are going into synapses. Neuron does a weighted summation on Inputs and their Corresponding Weights.

$$Input \times Weights = Sumed\ Output$$

Finally, neuron fires depending on whether the value reached a certain value or not. We can define this part as the Thresholding function or Activation function. We can use MTJ and HM structures to access their switching probability. As we mentioned before, switching probability of MTJ with HM is similar to the sigmoid function (see fig.3 (b)). And we called this Stochastic Sigmoid.

To better summarize, we have two operations:

- **Weighted Summation:** we can use crossbar of MTJs to calculate this operation.

- **Activation Function:** MTJ with HM can represent a stochastic sigmoid.

Based on the works in [3], Biological Spiking Neuron and MTJ Spiking neurons are shown in Fig. 3 (a). We can observe that the dynamics of MTJ neurons are described by the LLGS equation. Because of the inherently thermal noise during the switching process, there is inherent stochasticity during the switching.

### III. STOCHASTIC NEURAL NETWORKS

We can study stochastic neural networks with two different approaches: Spiking Neural Networks, Stochastic behavior of ANNs on Deterministic computing systems.

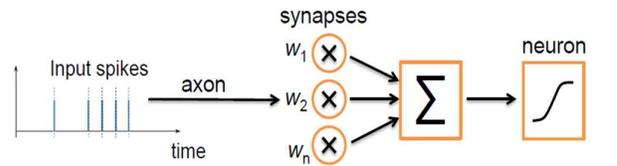

Fig.4.

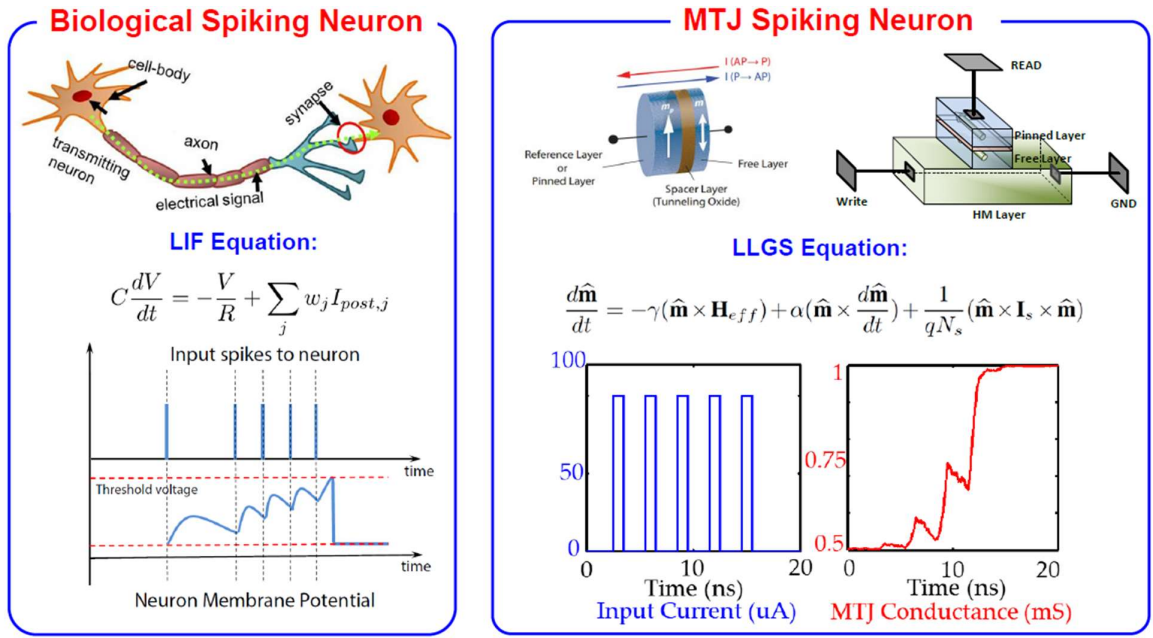

Fig.5. Dynamics of Biological Neurons and MTJ Spiking Neurons [3].

- **Spiking neural networks:**

In this approach, we design our neural network based on MTJ with HM structures. As previously discussed, MTJ can represent both neurons and synapses. We can achieve Computing-in-memory architecture using MTJs as neurons and synapses.

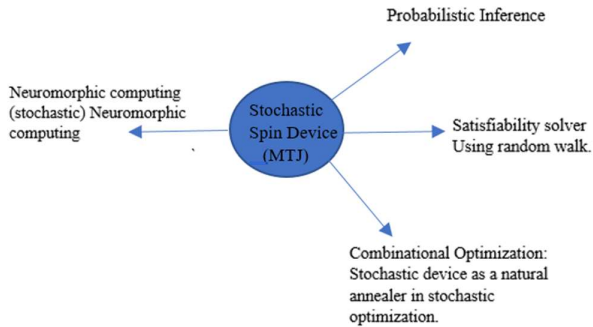

- **Stochastic Behavior of ANNs on Deterministic Computing Systems:**

Since most of the hardware and device technology that we use nowadays is based on deterministic computing and corresponding CMOS technology, it is important to introduce stochastic computing techniques implemented on deterministic platforms. For some specific applications, changing the device technology from CMOS to MTJ or even a combination of both might be unrealistic. One of the many methods that we can use to accelerate ANNs (DNN, CNN, RNN) is to use Stochastic Gradient Descent [4].

During the last decade, the data sizes have grown faster than the speed of processors. The computational complexity of learning algorithm becomes the critical limiting factor when one envisions very large datasets. to minimize the empirical risk $E_n(f_\omega)$ using gradient descent (GD). Each iteration updates the weights w on the basis of the gradient of $E_n(f_\omega)$,

$$w_{t+1} = w_t - \gamma \frac{1}{n} \sum_{i=1}^{n} \nabla_w Q(z_i, w_t),$$

where $\gamma$ is an adequately chosen gain. Each example z is a pair (x, y) composed of an arbitrary input x and a scalar output y. We consider a loss function $l(\hat{y}, y)$ that measures the cost of predicting $\hat{y}$ when the actual answer is y, and we choose a family F of functions $f_w(x)$ parametrized by a weight vector w. We seek the function $f \epsilon F$ that minimizes the loss $Q(z, w) = l(f_w(x), y)$ averaged on the examples [4].

The stochastic gradient descent (SGD) algorithm is a drastic simplification. Instead of computing the gradient of $E_n(f_\omega)$ exactly, each iteration estimates this gradient on the basis of a single randomly picked example $z_t$:

$$w_{t+1} = w_t - \gamma_t \nabla_w Q(z_t, w_t)$$

The stochastic process { $w_t$, t=1, . . . } depends on the examples randomly picked at each iteration. It is hoped that behaves like its expectation despite the noise introduced by this simplified procedure. Since the stochastic algorithm does not need to remember which examples were visited during the previous iterations, it can process examples on the fly in a deployed system. In such a situation, the stochastic gradient descent directly optimizes the expected risk, since the examples are randomly drawn from the ground truth distribution [4]. To better summarize the Stochastic Gradient Descent, we now compare the GD to SGD.

Stochastic gradient descent is an optimization algorithm for NNs, which dramatically speeds up the gradient descent. SGD is a stochastic approximation of GD. Meaning in SGD; we replace the actual gradient calculations for the entire set by an estimate.

In learning with GD method, for every sample we need to follow four steps:

- Follow the feedforward network
- Calculate the error
- Backpropagate the error
- Calculate the Gradient

GD gets slow for networks with a higher number of elements and samples. It is unsatisfactory to use GD for training large NNs. The reason for this is that for a training set of X, we need to take all of the samples into consideration, and this process is very time-consuming, especially for large networks. The solution is to take SGD as an alternative. Instead of looking at all data set,

- Take one sample
- Perform backpropagation, it is important to randomize the data set first. This is crucial for being stochastic.
- Calculate the gradient

In this method, at every sample, we are changing the behavior of our network. There is another alternative for GD and SGD, which is called minibatch gradient descent.

A batch is a group of samples used for the training and calculating gradient. In minibatch gradient descent, instead of taking the whole data set, we choose one mini-batch (one subset of data) at a time, and then we calculate the gradient. Fig. 6 better illustrates this method.

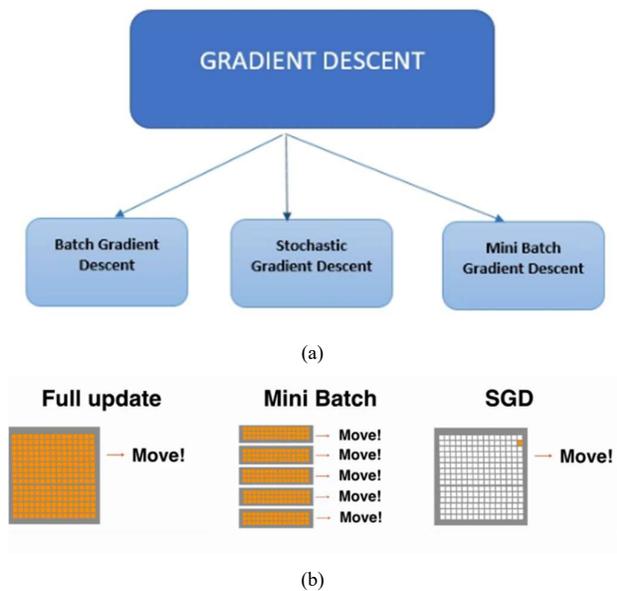

(a)

(b)

Fig.6 a) Variations of Gradient descent to optimize the performance of ANNs, b) difference in data size in each iteration for training.

**GD:**
- Needs to collect lots of feedback before making adjustments, in the plus side it requires fewer adjustment steps.
- With more complex and costly hardware, it could be processed parallelly.

**SGD:**
- Many small adjustments are required, in the other hand it spends less time collecting feedback between adjustments.
- It is not completely a parallel process, because we need feedback from one iteration to proceed with the next iteration.
- Preferable over GD if the full population is very large.

IV. STOCHASTIC NEURAL SUCSECIVE CANCELATION DECODERS FOR POLAR CODES

I Error-correction codes (ECC) or channel codes allow correcting errors in a data sequence without the need to refer to the original source of information. By allowing reliable data transmission and storage despite the inevitable noise that perturbs physical signals, they dramatically reduce the cost of computer systems, and as such play a central role in enabling the digital revolution that we witness today [1]. In this study, we focus on an important type of code called Polar Codes.

Polar Codes represent a class of error-correcting codes that is mathematically proven to achieve channel capacity with low complexity encoding and decoding algorithms [5]. The original Successive Cancellation decoder (SC) is too slow, and has a low throughput because of SC's bit-by-bit approach, meaning it decodes one bit and uses the previously estimated bit to the decode the next bit. The traditional successive-cancellation (SC) decoder suffers from very high latency, particularly at these long block-lengths, and the corresponding decoder algorithm has high complexity [6].

As we mentioned before, there are two different methods to implement Stochastic neural networks.

- Accelerating the designed ANN (RNN) based SC decoder, with stochastic computing techniques such as stochastic gradient descent or minibatch gradient descent.

- Spiking Neural Network (SNN) based SC decoder. This part mainly focuses on fabricating spiking neural networks using MTJ with HM devices. The specific configuration of SNN is required for SC Decoders.

As a future work we can focus on the device-level fabrication of SNNs and configuration of SNN layers in a way that it could be used for SC decoders.

## V. Conclusion

In this paper, we first compared the function of biological neurons and the corresponding Artificial models. As we deduced previously, ANNs are inefficient when it comes down to power consumption and processing power. To address the solution, we reviewed spintronic approaches in designing a stochastic computing platform that is capable of implementing neural networks more efficiently. This study led to the formation of Spiking neural networks designed by utilizing Magnetic Tunnel Junction devices and their optimized versions consisting of heavy metals (Platinum, or Tantalum). We proposed a stochastic computing technique (Stochastic Gradient Descent) to accelerate the ANN without changing the underlying deterministic computing architecture. As final words, we can conclude that real-life fabrication of MTJ devices will improve the overall efficiency of next-generation computing platforms (Neuromorphic Chips, Spiking Neural Networks), because it will bring us closer to mimic the function of human brain's nervous system.